\documentclass{article}

 \usepackage[main,final]{neurips_2025}
\usepackage[utf8]{inputenc} % allow utf-8 input
\usepackage[T1]{fontenc}    % use 8-bit T1 fonts
\usepackage{hyperref}       % hyperlinks
\usepackage{url}            % simple URL typesetting
\usepackage{booktabs}       % professional-quality tables
\usepackage{amsfonts}       % blackboard math symbols
\usepackage{nicefrac}       % compact symbols for 1/2, etc.
\usepackage{microtype}      % microtypography
\usepackage{xcolor}         % colors
\usepackage{amsmath}
\usepackage{amsmath} % assumes amsmath package installed
\usepackage{amssymb}  % assumes amsmath package installed
\usepackage{algorithm, algpseudocode}
\usepackage{todonotes}
\usepackage{subcaption}
\usepackage{multirow} 
\usepackage{tikz}
\usetikzlibrary{calc}
\usepackage{wrapfig}

\definecolor{tabblue}{HTML}{1F77B4}
\definecolor{tabred}{HTML}{D62728}

\title{First-order Sobolev Reinforcement Learning}
\author{%
  Fabian Schramm\thanks{Correspondence to fabian.schramm@inria.fr} \\
  Inria - ENS Paris \\
  PSL Research University
  \And
  Nicolas Perrin-Gilbert \\
  ISIR, CNRS \\
  Sorbonne Universit\'e \\
  \And
  Justin Carpentier \\
  Inria - ENS Paris \\
  PSL Research University
}

\begin{document}

\maketitle

\begin{abstract}
We propose a refinement of temporal-difference learning that enforces first-order Bellman consistency: the learned value function is trained to match not only the Bellman targets in value but also their derivatives with respect to states and actions.
By differentiating the Bellman backup through differentiable dynamics, we obtain analytically consistent gradient targets. Incorporating these into the critic objective using a Sobolev-type loss encourages the critic to align with both the value and local geometry of the target function.
This first-order TD matching principle can be seamlessly integrated into existing algorithms, such as Q-learning or actor–critic methods (e.g., DDPG, SAC), potentially leading to faster critic convergence and more stable policy gradients without altering their overall structure.
\end{abstract}

\section{Introduction}
Differentiable physics simulators~\cite{freeman2021brax,werling2021fast,lelidec2021differentiable} enable access to gradients of dynamics and rewards, yet standard model-based RL largely ignores this type of information, which might help to converge faster towards optimal solutions. 
In this work, we propose \emph{First-Order Sobolev Reinforcement Learning}, which enforces first-order Bellman consistency by matching both values and derivatives of the Bellman target via the chain rule through differentiable dynamics.
We augment temporal-difference learning with gradient-matching terms so that \(Q_\phi(s,a)\) aligns with both the value and the state/action-derivative targets implied by the Bellman backup through the simulator and, when applicable, the target policy. This brings gradient information into the critic in a principled way, rather than relying on implicit derivatives of a value-only fit.
Value-only TD regression does not constrain the critic’s derivatives: neural networks can match targets while exhibiting arbitrary local gradients,
which in turn might undermine gradient-based control \cite{czarnecki2017sobolev}. In actor–critic methods, the policy update depends directly on \(\nabla_a Q(s,a)\) 
\cite{silver2014deterministic, sac18} and can therefore benefit from improved critic’s action-gradients.
The resulting benefits are twofold. In Q-learning, matching Bellman-consistent derivatives accelerates learning of \(Q\) and improves greedy policies. In actor–critic, the more reliable \(\nabla_a Q\) yields better policy updates and supports first-order step selection (e.g., line search or trust-region safeguards), since the local Taylor model \(Q(s,a+\Delta a) \approx Q(s,a) + \nabla_a Q(s,a)^\top \Delta a\) is more trustworthy.
We derive Bellman-consistent gradient targets by differentiating the backup through differentiable dynamics, introduce a simple critic objective that jointly matches values and gradients, and show how to use it in Q-learning and actor–critic methods. 
A 1D toy study illustrates the effect on critic accuracy and greedy control, and we extend it to continuous-control tasks with differentiable simulators.

\section{Background}
We consider continuous control with deterministic, differentiable dynamics \( s' = f(s,a) \) and differentiable reward \( r(s,a) \). A parametric critic \( Q_\phi(s,a) \) and, for actor–critic, a deterministic \( \mu_\theta(s) \) or stochastic  \( \pi_\theta(s) \) policy  are trained off-policy with target networks.

In Q-learning, the temporal-difference target \(y\) and critic loss \(L_Q\) are given by:
\[
y_{\max}(s,a) = r(s,a) + \gamma \max_{a'} Q_{\phi_{\text{targ}}}(f(s,a), a'), 
\qquad
L^{\text{QL}}_{Q}(\phi) = \mathbb{E}\big[(Q_\phi(s,a) - y_{\max}(s,a))^2\big].
\]
In deterministic actor–critic (e.g., DDPG), the target uses the target actor:
\[
y_{\mu}(s,a) = r(s,a) + \gamma\, Q_{\phi_{\text{targ}}}\big(f(s,a),\, \mu_{\theta_{\text{targ}}}(f(s,a))\big),
\qquad
L^{\text{AC}}_{Q}(\phi) = \mathbb{E}\big[(Q_\phi(s,a) - y_{\mu}(s,a))^2\big].
\]
The actor maximizes \( J(\theta) = \mathbb{E}_s[Q_\phi(s,\mu_\theta(s))] \) with the deterministic policy gradient
\[
\nabla_\theta J(\theta) = \mathbb{E}_s\!\left[\nabla_a Q_\phi(s,a)\big|_{a=\mu_\theta(s)}\, \nabla_\theta \mu_\theta(s)\right],
\]
hence accurate \( \nabla_a Q_\phi \) is essential for stable and efficient policy improvement.
For stochastic policies, as in SAC, we optimize an entropy-regularized objective
\(
J_{\text{SAC}}(\theta)
= \mathbb{E}_{s\sim\mathcal{D},\, a\sim\pi_\theta}\!
\left[ Q_\phi(s,a) - \alpha \log \pi_\theta(a|s) \right].
\)
The critic reads
\[
y_{\text{SAC}}(s,a) = r(s,a) 
+ \gamma\, \mathbb{E}_{a'\sim\pi_{\theta_{\text{targ}}}}\!
\left[ Q_{\phi_{\text{targ}}}(f(s,a), a') - \alpha \log \pi_{\theta_{\text{targ}}}(a'|f(s,a)) \right].
\]
Gradients w.r.t. \(s\) and \(a\) propagate through this target analogously to the deterministic case, including the derivative of the entropy term when computing \(\nabla_a y_{\text{SAC}}\).

\section{Methodology}

We leverage differentiable simulators to enforce \emph{first-order Bellman consistency} in the critic.
For actor–critic targets, set \( a' = \mu_{\theta_{\text{targ}}}(s') \); for max-target Q-learning, set \( a' = \arg\max_{b} Q_{\phi_{\text{targ}}}(s',b) \) and treat \( a' \) as a stop-gradient variable.

\subsection{First-order Bellman targets}
\label{subsec:bellman-targets}
We define the Bellman target \( y(s,a) \in \{ y_{\mu}(s,a),\, y_{\max}(s,a) \} \). Differentiating the backup via the chain rule yields gradients that are consistent with the Bellman operator:
\begin{align*}
\nabla_s y &= \nabla_s r 
+ \gamma\!\left[ 
\nabla_{s'} Q_{\phi_{\text{targ}}}(s',a')\, \tfrac{\partial f}{\partial s}
+ \nabla_{a'} Q_{\phi_{\text{targ}}}(s',a')\, \tfrac{\partial a'}{\partial s'}\, \tfrac{\partial f}{\partial s}
\right], \\
\nabla_a y &= \nabla_a r 
+ \gamma\!\left[ 
\nabla_{s'} Q_{\phi_{\text{targ}}}(s',a')\, \tfrac{\partial f}{\partial a}
+ \nabla_{a'} Q_{\phi_{\text{targ}}}(s',a')\, \tfrac{\partial a'}{\partial s'}\, \tfrac{\partial f}{\partial a}
\right].
\end{align*}
For actor–critic methods, \( \tfrac{\partial a'}{\partial s'} = \tfrac{\partial \mu_{\theta_{\text{targ}}}}{\partial s'} \); in max-target Q-learning, we set \( \tfrac{\partial a'}{\partial s'} = 0 \) to respect the stop-gradient. For SAC, an additional term accounts for the derivative of the entropy term.

\subsection{Sobolev critic}
We train the critic using a Sobolev-type objective that matches both values and gradients:
\[
L_Q(\phi) = 
\mathbb{E}\!\left[
(Q_\phi - y)^2
+ \lambda_s \|\nabla_s Q_\phi - \nabla_s y\|^2
+ \lambda_a \|\nabla_a Q_\phi - \nabla_a y\|^2
\right].
\]
This enforces first-order Bellman consistency and improves the local Taylor approximation of \( Q_\phi \).
In actor–critic methods, the actor update remains
\(
\nabla_\theta J(\theta) = 
\mathbb{E}\!\left[ \nabla_a Q_\phi(s,a)\, \nabla_\theta \mu_\theta(s) \right],
\)
but the improved \( \nabla_a Q_\phi \) yields more reliable policy gradients and supports first-order step selection.

\subsection{Algorithm and practicalities}

We provide the full algorithm in Alg.~\ref{alg:1}. Training requires simulator Jacobians \( \tfrac{\partial f}{\partial s}, \tfrac{\partial f}{\partial a} \), reward gradients \( \nabla_s r, \nabla_a r \), and autograd for \( \nabla_{s'} Q_{\phi_{\text{targ}}}, \nabla_{a'} Q_{\phi_{\text{targ}}} \), as well as either \( \tfrac{\partial \mu_{\theta_{\text{targ}}}}{\partial s'} \) (actor–critic) or a stop-gradient through \( a' \) (max-target Q-learning).
The simulator Jacobians and reward gradients can be stored in an extended replay buffer after rollout.
Training stability can benefit from Polyak averaging of target networks, stopping gradients through all target-network paths, and using moderate coefficients \( \lambda_s,\lambda_a \) (optionally warmed up from zero). 
The same first-order Bellman gradient consistency applies whether the target $y$ is deterministic or entropy-regularized (SAC).

\begin{algorithm}[H]
\small
\caption{First-Order Sobolev RL (Q-learning or Actor–Critic)}
\label{alg:1}
\begin{algorithmic}[1]
\State Initialize critic \(Q_\phi\); if actor–critic, initialize actor \(\mu_\theta\)
\State Initialize target networks \(Q_{\phi_{\text{targ}}}\leftarrow Q_\phi\); if actor–critic, \(\mu_{\theta_{\text{targ}}}\leftarrow \mu_\theta\)
\Repeat
  \State Collect transitions \((s,a,r,s'=f(s,a))\) off-policy and store in replay buffer
  \State Optionally store simulator Jacobians \((\tfrac{\partial f}{\partial s}, \tfrac{\partial f}{\partial a})\) and reward gradients \((\nabla_s r, \nabla_a r)\) in the buffer
  \State Sample a minibatch \((s_i,a_i,r_i,s_i')\) (and cached derivatives, if available)
  \For{each \(i\)}
    \State If actor–critic: \(a_i'=\mu_{\theta_{\text{targ}}}(s_i')\); else: \(a_i'=\arg\max_{b} Q_{\phi_{\text{targ}}}(s_i',b)\) with stop-gradient
    \State \(y_i = r_i + \gamma\, Q_{\phi_{\text{targ}}}(s_i',a_i')\)
    \State Compute \(\nabla_s y_i,\nabla_a y_i\) via the chain rule through \(f\) (and \(\mu_{\theta_{\text{targ}}}\) if applicable)
  \EndFor
  \State Update critic by minimizing
  \[
  \tfrac{1}{B}\sum_i \big[(Q_\phi(s_i,a_i)-y_i)^2 + 
  \lambda_s \|\nabla_s Q_\phi-\nabla_s y_i\|^2 + 
  \lambda_a \|\nabla_a Q_\phi-\nabla_a y_i\|^2\big]
  \]
  \State If actor–critic: update \(\mu_\theta\) with 
  \( \nabla_\theta J(\theta) = \tfrac{1}{B}\sum_i \nabla_a Q_\phi(s_i,a)\rvert_{a=\mu_\theta(s_i)} \nabla_\theta \mu_\theta(s_i) \)
  \State Optionally Polyak-update targets
\Until{convergence}
\end{algorithmic}
\end{algorithm}

\section{Experiments}

\noindent\textbf{1D toy problem.} 
We study a simple control problem to isolate the effect of Sobolev training on the critic. The state and action spaces are \(s,a\in[-1,1]\), dynamics are \(s' = f(s,a)=a\), the reward is \(r(s,a)=0.2\,a - (a-s)^2\), and the discount factor is \(\gamma=0.9\). 
We use Q-learning with the max target
\[
y(s,a)=r(s,a)+\gamma \max_{a'\in[-1,1]} Q_{\phi_{\mathrm{targ}}}(s',a'), \qquad s'=a,
\]
where the maximization over \(a'\) is approximated on a dense grid for training and evaluation.
This system admits a closed-form solution for the optimal value, Q-function, and policy.  The unconstrained optimum is \(a^*(s)=s+1\), which becomes piecewise after clipping to the action bounds, and allows analytical computation of \(Q^*\) and \(V^*\). These analytic solutions serve as ground truth for quantitative evaluation and for visual comparison.
We compare two parameterizations for \(Q_\phi\). First, a quadratic model \(Q_\phi(s,a)=\theta_0 + \theta_1 s + \theta_2 a + \theta_3 s^2 + \theta_4 sa + \theta_5 a^2\) with only six parameters and second, a neural critic implemented as a three-layer MLP with 128 units per layer and leaky ReLU activations. Both are trained using Adam with a learning rate of \(10^{-4}\) and a batch size of 50. At each iteration, we sample \((s,a)\) uniformly from \([-1,1]^2\), compute \(s'=a\) and \(r(s,a)\), and evaluate the \(\max_{a'}\) target on a 100-point grid. 
The Sobolev critic minimizes
\[
L_Q(\phi)=\mathbb{E}\big[(Q_\phi - y)^2 + \lambda_s \|\nabla_s Q_\phi - \nabla_s y\|^2 + \lambda_a \|\nabla_a Q_\phi - \nabla_a y\|^2\big],
\]
with \(\lambda_s=\lambda_a=1\), while the value-only baseline uses \(\lambda_s=\lambda_a=0\).

\begin{wrapfigure}[10]{r}{0.6\textwidth}
\vspace{-1em}
\centering
\captionof{table}{Quantitative results (avg. over 5 seeds) on the 1D control problem. 
Reported are mean-squared errors (MSE) with respect to \(Q^*\), 
\(\nabla_a Q^*\), and the induced policy.}
\label{tab:toy}
\vspace{0.3em}
{\tiny
\begin{tabular}{lcccc}
\toprule
Model & Method & \(Q^*\) MSE & \(\nabla_a Q^*\) MSE & Policy error \\
\midrule
\multirow{2}{*}{Quadratic} 
& Baseline & $4.10{\scalebox{0.7}{$\pm0.8\times 10^{-2}$}}$ & $1.93{\scalebox{0.7}{$\pm0.4\times 10^{-1}$}}$ & $3.80{\scalebox{0.7}{$\pm0.1\times 10^{-2}$}}$ \\
& Sobolev  & $1.05{\scalebox{0.7}{$\pm0.2\times 10^{-2}$}}$ & $3.16{\scalebox{0.7}{$\pm0.8\times 10^{-2}$}}$ & $5.03{\scalebox{0.7}{$\pm0.4\times 10^{-3}$}}$ \\
\midrule
\multirow{2}{*}{MLP} 
& Baseline & $4.45{\scalebox{0.7}{$\pm2.8\times 10^{-2}$}}$ & $4.33{\scalebox{0.7}{$\pm1.6\times 10^{-1}$}}$ & $1.67{\scalebox{0.7}{$\pm1.1\times 10^{-2}$}}$ \\
& Sobolev  & $1.25{\scalebox{0.7}{$\pm2.6\times 10^{-2}$}}$ & $2.14{\scalebox{0.7}{$\pm2.2\times 10^{-2}$}}$ & $1.35{\scalebox{0.7}{$\pm2.0\times 10^{-3}$}}$ \\
\bottomrule
\end{tabular}
}
\vspace{-1em}
\end{wrapfigure}
\begin{figure}[tb]
    \centering
    \includegraphics[width=\linewidth]{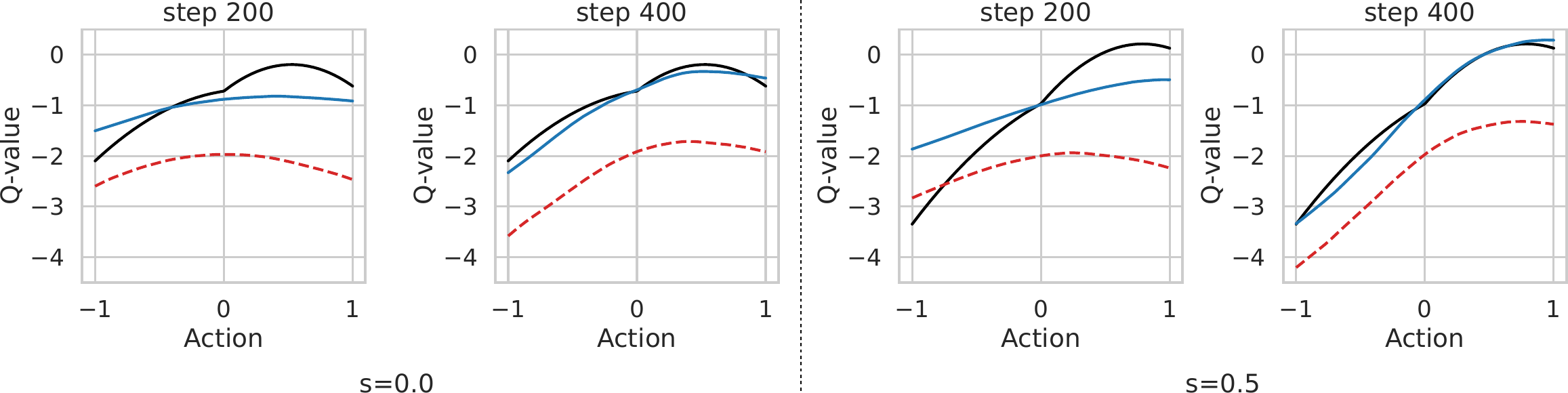}
    \caption{\textbf{Comparison of $Q$-function slices:} ground-truth (black), \textcolor{tabblue}{Sobolev Q-learning}, and dashed \textcolor{tabred}{default Q-learning} after 200 and 400 training step for two different states $s=0.0$ and $s=0.5$.}
    \label{fig:q_slices}
    \vspace{-1.6em}
\end{figure}

We evaluate performance in terms of mean-squared error (MSE) to the analytic \(Q^*\), the gradient MSE with respect to \(\nabla_a Q^*\), and the return of the greedy policy \(\pi(s)=\arg\max_a Q_\phi(s,a)\), estimated via Monte Carlo rollouts. Fig.~\ref{fig:q_slices} visualizes learned \(Q\)-slices against \(Q^*\), which reveals how Sobolev training accelerates the convergence of both function shape and derivatives. Tab.~\ref{tab:toy} shows the quantitative results: Sobolev training reduces both value and gradient errors for both model classes.
Enforcing first-order Bellman consistency enhances the geometric accuracy of the learned \(Q\)-function and quality of the induced policy.

\noindent\textbf{Continuous control task using differentiable simulator.} 
To assess the applicability beyond toy settings, we train SAC with and without Sobolev critics on a differentiable MuJoCo Ant task using \texttt{Rewarped}~\cite{xing2025stabilizing}. The simulator provides differentiable dynamics \( f \) and reward gradients \( \nabla_s r, \nabla_a r \), allowing computation of first-order Bellman targets as in Section~\ref{subsec:bellman-targets}. Both agents share identical network architectures and hyperparameters; the only difference is the additional gradient-matching terms in the critic loss.
Training uses learning rate \(5\times10^{-3}\), target smoothing \(\tau=0.99\), batch size 2048, and discount \(\gamma=0.99\). 
For Sobolev critics, \(\lambda_s=1;\lambda_a=0.1\), gradients through target networks are stopped and results are averaged over 5 seeds.

\begin{wrapfigure}[10]{r}{0.6\textwidth}
\vspace{-1em}
\centering
\begin{subfigure}[t]{0.48\linewidth}
    \centering
    \includegraphics[width=\linewidth]{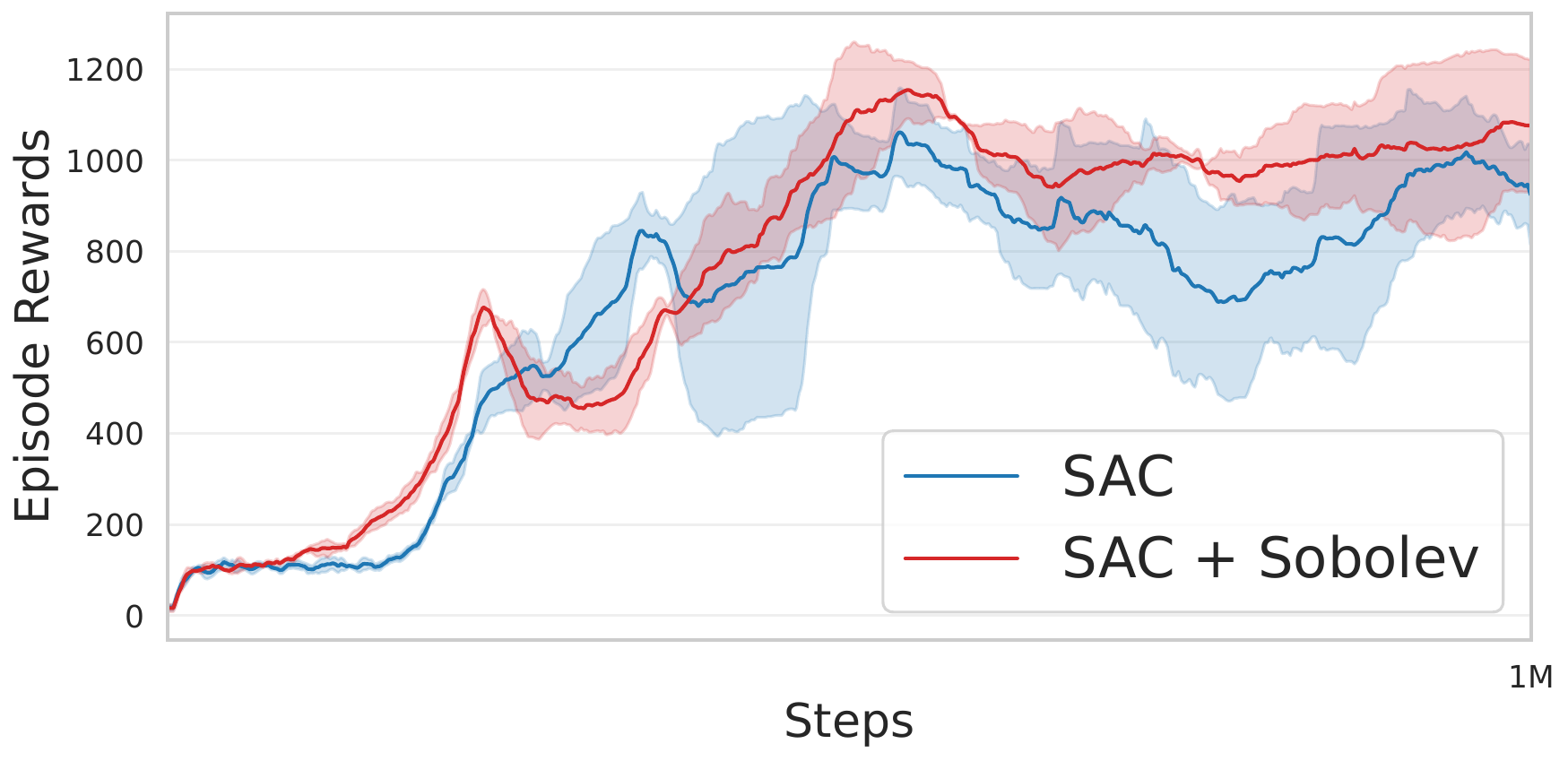}
    \caption{Episode Reward.}
    \label{fig:baseline}
\end{subfigure}%
\hfill
\begin{subfigure}[t]{0.48\linewidth}
    \centering
    \includegraphics[width=\linewidth]{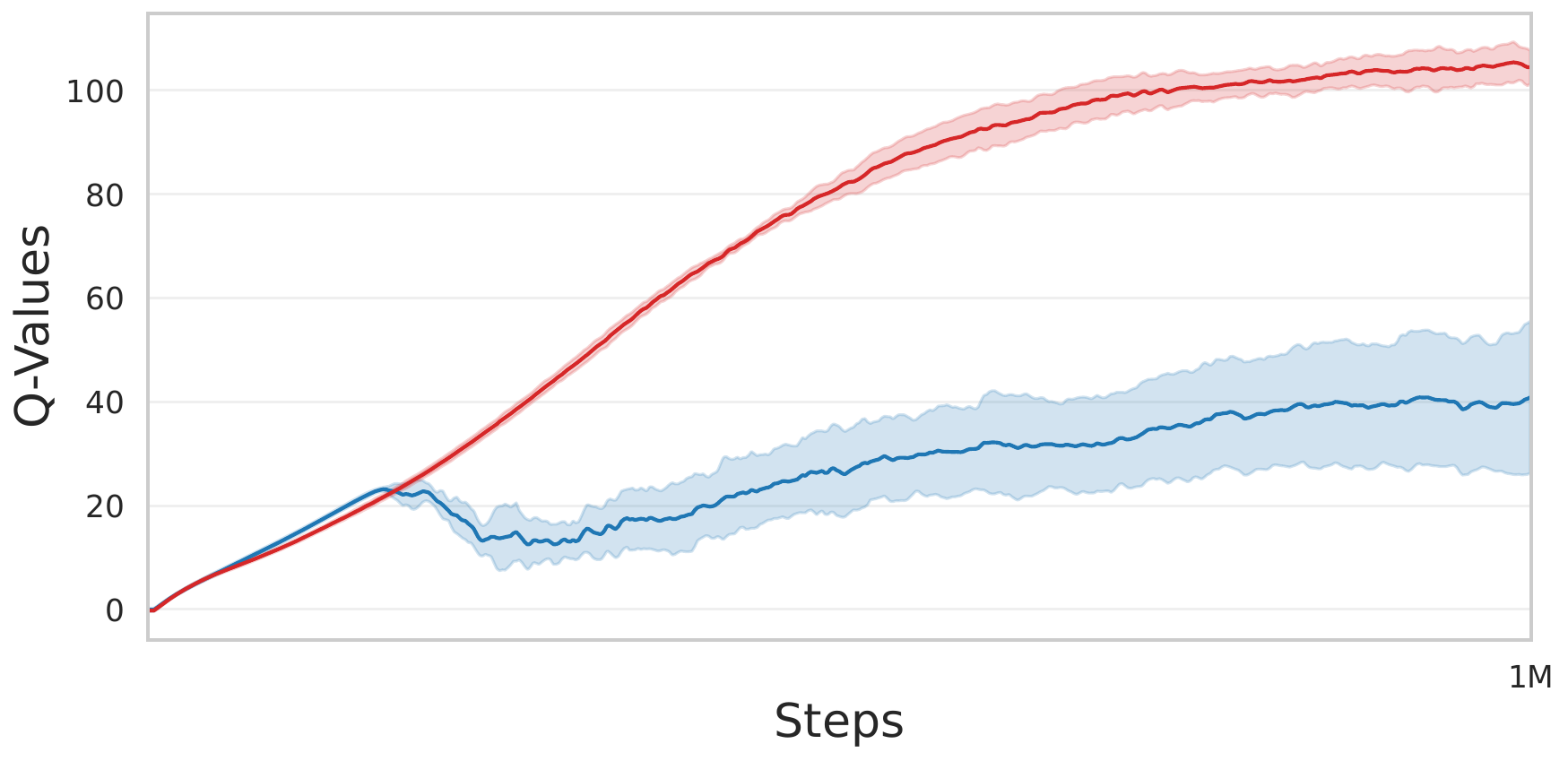}
    \caption{Mean Q-Values.}
    \label{fig:sobolev}
\end{subfigure}
\caption{\textbf{Comparison on MuJoCo Ant}.}
\label{fig:ant}
\vspace{-0.5em}
\end{wrapfigure}
The Sobolev critic leads to faster and smoother initial learning compared to the standard SAC baseline. Q-values for a held-out set of 1000 states converge earlier and fluctuate less, suggesting more stable Bellman updates (see Fig.~\ref{fig:ant}). However, overall performance remains comparable, indicating that while the modified critic improves early dynamics, it does not yet translate into consistently higher returns. Future work should explore adaptive scheduling of the first-order weighting and inclusion of a line search to balance gradient and value consistency better.

\section{Related Work}
Recent efforts have explored incorporating first-order information directly into reinforcement learning updates by differentiating through dynamics. SHAC~\cite{xu2021accelerated} and PODS~\cite{mora2021pods} exploit differentiable simulators to compute analytic gradients of returns, enabling efficient on-policy optimization in smooth environments. However, such fully differentiable on-policy approaches often suffer from exploding or vanishing gradients over long horizons, and become unstable when encountering non-smooth transitions or contact dynamics~\cite{georgiev2024adaptivehorizonactorcriticpolicy}. 

Off-policy methods, by contrast, are generally more sample-efficient~\cite{Fujimoto2018AddressingFA,lillicrap2015continuous,sac18} and can reuse experience across updates, but so far have primarily relied on value matching rather than enforcing gradient consistency. Our work aims to bridge this gap by introducing first-order Bellman consistency, a local, single-step gradient matching principle that leverages simulator Jacobians without requiring complete trajectory differentiation. This eliminates gradient accumulation issues, remains compatible with standard off-policy algorithms, and provides a stable route toward first-order reinforcement learning in complex continuous-control domains.

\section{Conclusion}
Enforcing first-order Bellman consistency yields critics that better align values and derivatives, improving local accuracy and gradient usefulness for control.
Preliminary results show smoother learning dynamics and more stable value estimates, suggesting that gradient consistency can complement standard value matching in off-policy RL.
This opens a promising direction for developing first-order off-policy methods that better exploit differentiable simulators, for instance, through adaptive weighting or line-search strategies to balance value and derivative objectives.

\section*{Acknowledgments}
This work has received support from the French government, managed by the National Research Agency, under the France 2030 program with the references Organic Robotics Program (PEPR O2R) and “PR[AI]RIE-PSAI” (ANR-23-IACL-0008).
This research was funded, in part, by l’Agence Nationale de la Recherche (ANR), projects RODEO (ANR-24-CE23-5886) and PEPR O2R – PI3 ASSISTMOV (ANR-22-EXOD-0004).
The European Union also supported this work through the ARTIFACT project (GA no.101165695) and the AGIMUS project (GA no.101070165).
The Paris Île-de-France Région also supported this work in the frame of the DIM AI4IDF.
Views and opinions expressed are those of the author(s) only and do not necessarily reflect those of the funding agencies.

{
\small

\bibliographystyle{plainnat}
\bibliography{references}

\end{document}